%% file: main.tex
\documentclass[conference]{IEEEtran}
\IEEEoverridecommandlockouts
\usepackage{cite}
\usepackage{amsmath,amssymb,amsfonts}
\usepackage{algorithmic}
\usepackage{graphicx}
\usepackage{textcomp}
\usepackage{xcolor}
\usepackage{multirow}
\usepackage{threeparttable}
\usepackage{subcaption}

\begin{document}

\title{
    %Early Event-Based Action Recognition with Two-Stream Gated Spiking Neural Networks\\
    %Who spikes first? Early Event-Based Action Recognition with High-Rate Two-Stream Spiking Neural Networks\\
    EEvAct: Early Event-Based Action Recognition with High-Rate Two-Stream Spiking Neural Networks
    %{
    %    \footnotesize 
    %    \textsuperscript{*}Note: Sub-titles are not captured for https://ieeexplore.ieee.org and should not be used
    %}
    \thanks{This project has been funded funded by the Bayerische Forschungsstiftung under the grant number AZ-1558-22.}
}

\author{\IEEEauthorblockN{1\textsuperscript{st} Neumeier Michael}
\IEEEauthorblockA{\textit{fortiss GmbH and} \\
\textit{Technical University Munich}\\
Munich, Germany \\
neumeier@fortiss.org}
\and
\IEEEauthorblockN{2\textsuperscript{nd} Lecomte Jules}
\IEEEauthorblockA{\textit{Neuromorphic Computing,} \\
\textit{fortiss GmbH}\\
Munich, Germany }
%0000-0002-7103-0843}
\and
\IEEEauthorblockN{3\textsuperscript{rd} Kazinski Nils}
\IEEEauthorblockA{\textit{Simi Reality Motion Systems GmbH} \\
Unterschleißheim, Germany 
}
\and
\IEEEauthorblockN{4\textsuperscript{th} Banik Soubarna}
\IEEEauthorblockA{\textit{Simi Reality Motion Systems GmbH} \\
Unterschleißheim, Germany 
}

\and
 \IEEEauthorblockN{5\textsuperscript{th} Li Bing}
 \IEEEauthorblockA{\textit{Digital Integrated Systems
Group,} \\
 \textit{University of Siegen}\\
 Siegen, Germany 
}

 \and
 \IEEEauthorblockN{6\textsuperscript{th} von Arnim Axel}
 \IEEEauthorblockA{\textit{Neuromorphic Computing,} \\
 \textit{fortiss GmbH}\\
 Munich, Germany }
 }

\maketitle

\input{sec/0_abstract}    
\input{sec/1_intro}

\input{sec/2_related_work}

\input{sec/3_methods}
\input{sec/4_experiments}

\input{sec/5_conclusion}
%{
%    \small
%    \bibliographystyle{ieeenat_fullname}
%    \bibliography{main}
%}

%\section*{References}

\bibliographystyle{IEEEtran}
\bibliography{main}

\end{document}

%% file: sec/0_abstract.tex
\begin{abstract}
Recognizing human activities early is crucial for the safety and responsiveness of human-robot and human-machine interfaces. Due to their high temporal resolution and low latency, event-based vision sensors are a perfect match for this early recognition demand. However, most existing processing approaches accumulate events to low-rate frames or space-time voxels which limits the early prediction capabilities. In contrast, spiking neural networks (SNNs) can process the events at a high-rate for early predictions, but most works still fall short on final accuracy. In this work, we introduce a high-rate two-stream SNN which closes this gap by outperforming previous work by 2\% in final accuracy on the large-scale THU EACT-50 dataset. We benchmark the SNNs within a novel early event-based recognition framework by reporting Top-1 and Top-5 recognition scores for growing observation time. Finally, we exemplify the impact of these methods on a real-world task of early action triggering for human motion capture in sports.
\end{abstract}


%% file: sec/1_intro.tex
\section{Introduction}
\label{sec:intro}

% Motivation:
% \begin{itemize}
%     \item Human action recognition as a use-case: AR/VR, surveillance, sign language, human-machine or human-robot interaction
%     \item Overview sensor modalities for action recognition
%     \item Requirements: always-on, low power, resource constraints
%     \item Early action recognition and anticipation
%     \item Event-based vision for action recognition
%     \item Different ways of processing event streams
%     \item Event-by-event processing with spiking neural networks
%     \item Ways of benchmarking SNN approaches
% \end{itemize}

Human action or activitity recognition (HAR) enables devices, machines or robots to understand human behavior. Hence, HAR is a crucial task in various application domains like AR/VR~\cite{EpicKitchens_Dataset}, surveillance~\cite{online_har}, autonomous driving~\cite{pedestrian_action}, human-robot interaction~\cite{early_event_manact} and sports~\cite{har_sports}.
In all these use-cases, sensors perceive the human activities and processors analyze the sensor data to interpret the human movements. Different sensors like inertial measurement units (IMUs)~\cite{imu_har}, Radar~\cite{radar_har} and RGB cameras~\cite{EpicKitchens_Dataset, UCF101_Dataset} are used for perception. State-of-the-art processing is nowadays performed using deep neural networks~\cite{Survey_act_ant}. Typical use-case requirements are always-on operating mode and low power consumption~\cite{streaming_action}. Next to these, the time until a human action is identified by the system is also a critical requirement. Accordingly, early action recognition and anticipation has evolved as a field of study and a benchmarking task~\cite{online_har, Survey_act_ant}. Therefore, algorithms are evaluated for partially observed sensor data (early recognition) or for the prediction of future actions (anticipation).

\begin{figure}[t]
\centering
\includegraphics[width=\columnwidth]{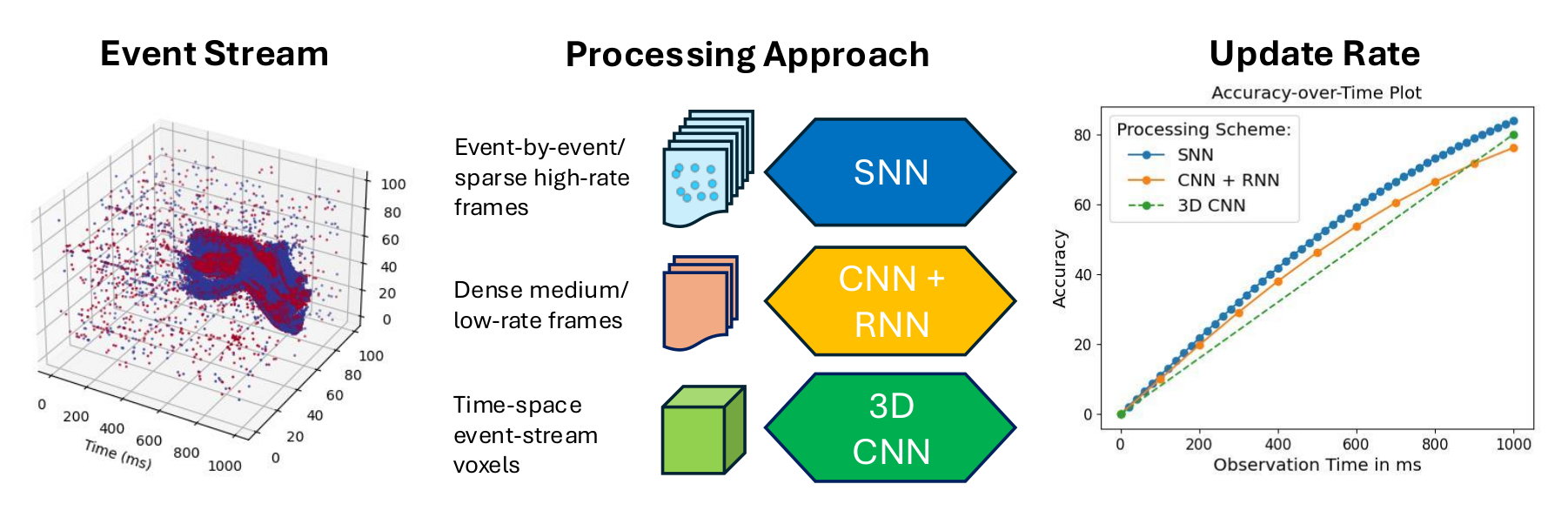}
\caption{Event stream processing: Event-by-event, high-rate SNNs, medium frame rate 2D CNNs plus RNN layers and voxel-based 3D CNNs with corresponding update rates.}
\label{fig:intro}
\end{figure}

Event-based vision sensors~\cite{event_survey} have also been proposed for human activity recognition. With their properties of low power consumption, data sparsity and high temporal resolution they are a good fit to the above-mentioned requirements~\cite{event_survey}. In Figure~\ref{fig:intro}, we show how event streams are typically processed. 3D CNNs~\cite{thu_eact_50} accumulate a space-time 3D cube of event voxels on which they make a prediction. 2D CNNs plus RNN layers~\cite{event_cnn_lstm} process medium or low-rate dense event frames. In comparison, the state-space models in \cite{event_ssm} and spiking neural networks (SNNs) are designed to process the streams event-by-event or accumulated in sparse, high-rate frames~\cite{event_survey}. They asynchronously integrate sparse events and emit sparse spikes if the accumulated information exceeds a threshold. This leads to a high-rate output prediction. Early works theoretically highlighted a computational advantage of SNNs over traditional artificial neural networks (ANNs)~\cite{maass_snn}. In these studies the events are in form of binary "1s", but recent works propose that they can also carry additional information. Novel neuromorphic processors are also capable of implementing both forms of events efficiently~\cite{loihi2}. These characteristics indicate a high potential of SNNs for event-based HAR.\\
However, SNNs still do not achieve similar accuracy as ANNs, particularly on large-scale datasets~\cite{thu_eact_50}. Moreover, the existing approaches are generally only evaluated for the final accuracy, not for early recognition on partially observed event streams.\\
In this work, we try to close this accuracy gap on the large-scale event-based human activity dataset THU EACT-50~\cite{thu_eact_50}. To achieve this, we propose a novel two-stream SNN architecture and heterogenous spiking neuron layers with trainable decay, feedback and gating parameters. We also formalize an early event-based action recognition benchmark and optimize the proposed SNNs therefore. We study the impact of different loss and readout options and conclude a novel readout layer which combines early approximate and long-term accurate recognition. Finally, we apply these findings to a real-world use-case in form of an event-based trigger system for human motion capture in sports.\\
We summarize our contributions as follows:
\begin{enumerate}
    \item We formalize an evaluation scheme for early event-based recognition which incorporates Top-K accuracy, output update rate and effective SynOps estimation
    \item We propose a novel two-stream SNN architecture with convolutional spiking neuron layers and event-based gated units. It achieves new state-of-the-art accuracy of 94.9\% on THU EACT-50, outperforming the 3D CNN from \cite{thu_eact_50} by more than 2\% while requiring only 20\% of parameters and less than 1\% of synaptic operations.
    \item We introduce a combined loss function and readout scheme which allows for 50\% Top-5 accuracy within the first 100ms of observation and close to state-of-the-art final Top-1 accuracy of 90\% on THU EACT-50.
\end{enumerate}

In section~\ref{sec:related_work} we analyze related work around early action recognition, event-based action recognition and recent advancements in the field of SNNs. Next, we introduce our early event-based recognition framework in section~\ref{sec:eear}. In section~\ref{sec:twosat} we propose the two-stream spiking neural network, followed by in-depth evaluation within the early event-based recognition framework in section~\ref{sec:experiments}. Finally, we apply the methods to a real-world sports use-case and draw our conclusion in section~\ref{sec:conclusion}.

% Intro picture: Comparison of event-stream processing with 3D CNN/Transformers, 2D CNNs and SNNs
% \begin{itemize}
%     \item left: 3x event-stream in 3D + indicate input sample
%     \item middle: networks
%     \item right: 3x accuracy vs. time plots
% \end{itemize}

%% file: sec/2_related_work.tex
\section{Related Work}
\label{sec:related_work}

\subsection{Early RGB-based Action Recognition}

%\begin{itemize}
%    \item DNNs for action recognition
%    \item Early action recognition
%    \item Action anticipation (short-term, long-term)
%    \item Way of benchmarking
%\end{itemize}

Using cameras for HAR offers the advantage over IMU-based HAR that no wearable or portable sensors are necessary. Hence, action recognition from RGB videos is a heavily studied field of research. Large-scale datasets for both third-person~\cite{UCF101_Dataset, HMDB_Dataset} and ego-centric first person action recognition~\cite{EpicKitchens_Dataset} are available for benchmarking. First deep learning approaches for RGB action recognition were based on 2D CNNs for per-frame feature extraction and RNN layers like LSTM units for temporal aggregation~\cite{CNN+LSTM}. In contrast to this iterative approach, 3D CNNs like C3D~\cite{C3D} evaluate the temporal sequence at once. Some approaches also use preprocessed representations like human joint positions~\cite{G3D} or optical flow~\cite{TwoPathCNN, I3D}. Two-Path networks~\cite{TwoPathCNN} apply 2D CNNs to both RGB frames and optical flow. I3D~\cite{I3D} networks add 3D convolutions on top of this. SlowFast~\cite{SlowFast} approaches have a coarse-temporal sampled 3D CNN and fine-temporal sampled 3D CNN path. Recent works also introduced transformer architectures which outperformed the CNN-based approaches~\cite{TransformerHAR}. Nevertheless, our work takes inspiration from CNN-based approaches, particularly the separation of spatial and temporal processing in a two-path architecture.\\
For the task of action recognition the goal is to classify the actions in a video based on full observation~\cite{Survey_act_ant}. This is extended to online action recognition by using untrimmed videos in real-time~\cite{online_har}. For streaming HAR, the runtime of the model is also taken into account~\cite{streaming_action}. Evaluation for partially observed videos is called early action recognition~\cite{Survey_act_ant}. Action anticipation also expands the evaluation to future, not yet observed activities~\cite{Survey_act_ant}. Probabilistic approaches, sensor fusion and knowledge distillation have been proposed as anticipation optimization methods~\cite{Survey_act_ant}. Recently, modern RNNs based on deep structured state-space models (SSMs) outperformed transformers on this task~\cite{Manta}.\\
For both early recognition and anticipation, benchmarking accuracy scores for different observation ratios is analyzed in a table or an accuracy-over-observation-ratio plot. For online evaluation the observation ratio is replaced with observation time~\cite{early_event}. In this work we formalize a similar benchmarking scheme for event-based action recognition and extend it to incorporate the output update frequency and the computational performance.

\subsection{Event-based Action Recognition}
%\begin{itemize}
%    \item Gesture recognition (DVSGesture)
%    \item Action recognition (Early datasets, THU EACT-50)
%    \item Way of processing $\xrightarrow{}$ gap: spiking leads to accuracy deficit
%    \item Way of benchmarking $\xrightarrow{}$ gap: not highlighting the power of SNNs
%\end{itemize}
Event-based vision sensors capture changes in luminosity in their field of view with high temporal resolution and high dynamic range~\cite{event_survey}. These properties make them a good fit for perceiving human movements and hence, for action recognition. Accordingly, multiple datasets have been recorded and published. DVSGesture~\cite{dvs_gesture} has around 1500 samples labeled for 11 upper-body gesture classes. Similar-sized event-based datasets have been introduced for whole-body action recognition~\cite{dvs_action, motion_snn}. In \cite{graph_nvs} event-based versions of the RGB datasets UCF101-DVS and HMDB51-DVS were provided. These were recorded with an event camera while displaying the RGB videos on a screen. This recording procedure might miss crucial properties of event-based vision sensors like high temporal resolution and dynamic range. In contrast to this, THU EACT-50~\cite{thu_eact_50} is a large-scale action recognition dataset directly recorded with an event camera.\\
To process the event streams, most approaches convert them into frames. The subsequent videos of event frames are processed with 2D CNNs plus recurrent layers~\cite{event_cnn_lstm}, 3D CNNs~\cite{thu_eact_50}, transformers~\cite{event_frame_transformer} or a combination of graph neural networks and 3D CNNs~\cite{graph_nvs}. These approaches trade using knowledge gained from RGB-based action recognition for the sparsity and high temporal resolution of raw event streams. In contrast, other approaches directly process events using point-based networks~\cite{event_point} or SSMs~\cite{event_ssm}. The SSM proposed in \cite{event_ssm} achieved close to state-of-the-art accuracy on the DVSGesture dataset. Spiking neural networks were also introduced to directly process events in a sparse and asynchronous manner~\cite{event_survey}. However, to train them on a spatiotemporal task like action recognition, events are still converted to frames~\cite{snn_deep_learning}. In this work, we investigate this trade-off and propose to shorten the duration of one frame as much as possible in order to foster event-by-event-like processing.\\
Most of the approaches simply report accuracy scores for the complete action sequence. Solely in \cite{early_event_manact}, the evaluation of their transformer model is performed for early recognition using accuracy-over-time plots. Building upon this evaluation idea, we introduce a formalized early recognition benchmark and run a comparative study for different SNN variants.

\subsection{Advanced Spiking Neural Networks}

%\begin{itemize}
%    \item SNN approaches for action recognition
%    \item Trainable decays $\xrightarrow{}$ PLIF
%    %\item trainable delays $\xrightarrow{}$ Learning delays % outlook
%    \item Bioinspiration: adaptive spiking neurons $\xrightarrow{}$ AdLIF
%    \item RNN-inspiration: Gating in spiking neurons $\xrightarrow{}$ LTC, SpikGRU, EGRU
%    \item TET loss (early audio)
%    \item NeuroBench
%\end{itemize}

Recent work proposed event-based action recognition as a task to understand the advantages of spike-based processing~\cite{dvs_gesture_chain}. They evaluate a spiking ResNet on chained gestures from the DVSGesture dataset~\cite{dvs_gesture}. Moreover, they also compare SNN with ANN and RNN architectures as well as different learning algorithms. 
Novel architectures for SNNs inspired by PointNets~\cite{spike_point}, 3D CNNs~\cite{spike_3d} or transformers~\cite{spike_har++}, are proposed and perform well for event-based action recognition. However, these architectures are not well suited for streaming early action recognition and efficient implementation on neuromorphic hardware like Intel Loihi 2~\cite{loihi2}.\\
A motion-based SNN for event-based action recognition is proposed in \cite{motion_snn}. In a follow-up work this approach was also benchmarked on the THU EACT-50 dataset, but with a notable drop in recognition accuracy compared to their SlowFast 3D CNN~\cite{thu_eact_50}. A two-stream SNN consisting of a spatial and a motion path is introduced in \cite{two_stream_snn}, but only evaluated on the small-scale event-based action recognition datasets. These two approaches operate in an iterative fashion and are better suited for streaming early action recognition using neuromorphic hardware. We extend these ideas with advanced, trainable spiking neuron units and towards smaller time bins for event-by-event like processing and early recognition. \\
These extensions are based on recent advancements in trainable spiking neuron units. Parametric LIF (PLIF)~\cite{plif} have a trainable decay factor per layer. Closer biologically-inspired neuron units based on adaptive LIF (adLIF)~\cite{neuronal_dynamics} have been used for small-scale classification and regression~\cite{adlif}. These units can express more complex behaviours like spike frequency adaptation and membrane potential oscillation. Besides, gating mechanisms were also added to spiking units. Examples are SpikGRUs~\cite{spikgru} and EGRUs~\cite{egru}. The latter can either emit binary spikes or multi-bit events.
We use the mentioned emerging spiking units on a larger scale within convolutional spiking layers and for a complex action recognition task.\\
Typically, the loss terms, to train SNNs on action recognition tasks, are based on spike rate or cross entropy (CE) loss on mean, maximum or last membrane potential value of the output neurons~\cite{snn_deep_learning}. Recent works extended this to sample the membrane potentials at multiple time-steps and calculate the average CE loss. This improved accuracy and enabled SNNs with less timesteps for RGB image classification~\cite{tet_loss}. A similar loss term was also used for early audio classification~\cite{early_audio_snn}. In our experiments we study the impact of these different loss terms on early recognition performance. Furthermore, we build upon the recently proposed NeuroBench~\cite{neurobench}, a benchmarking framework for neuromorphic computing and propose a possible extension to early event-based action recognition.
% Neurobench

%% file: sec/3_methods.tex
\begin{figure*}[h]
\centering
\includegraphics[clip,width=0.8\textwidth]{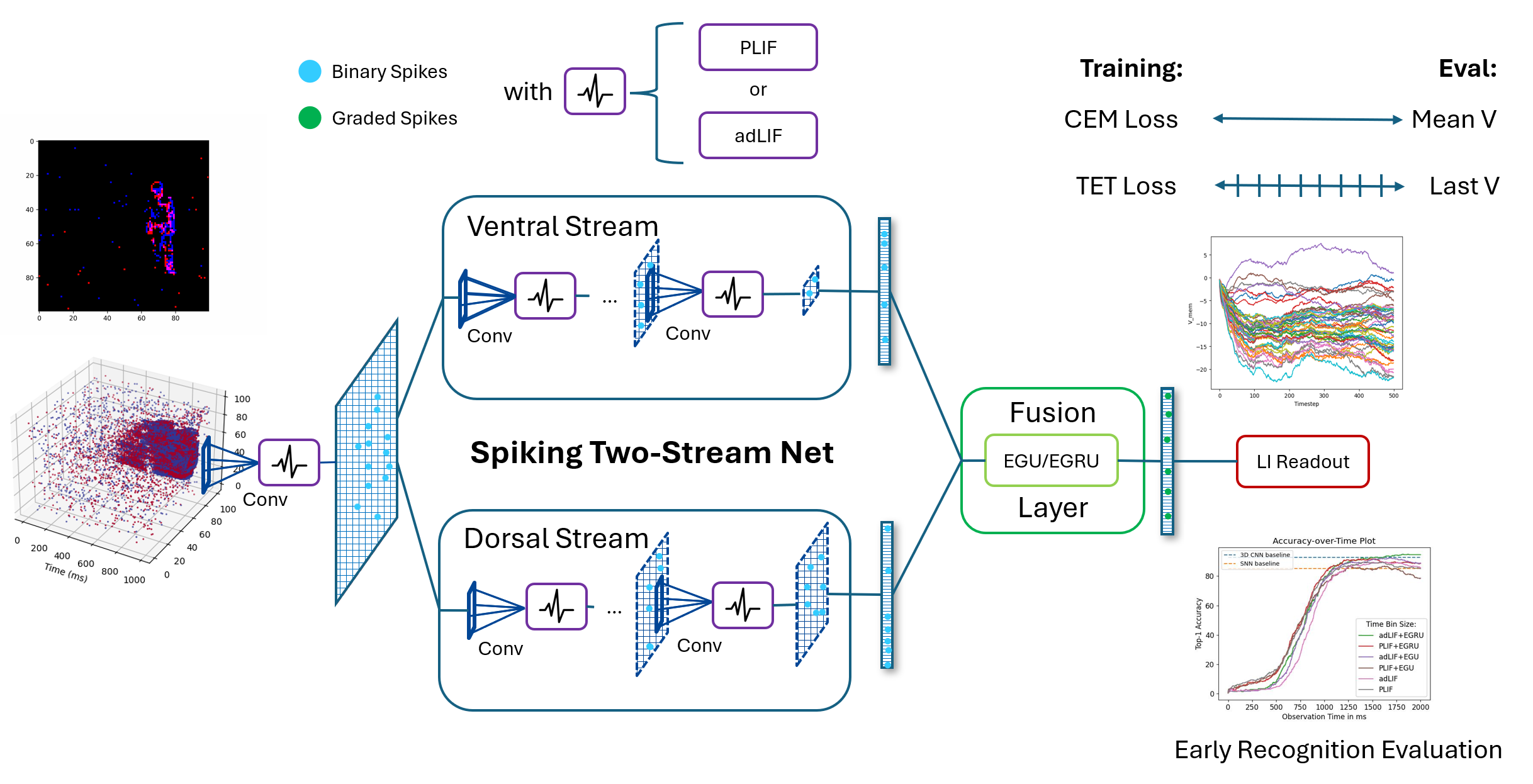}
\caption{Spiking Two-Stream Net: a common spiking convolution layer, parallel ventral and dorsal streams, an event-based gated fusion layer and LI readout. We also visualize the two loss/readout principles and early recognition evaluation scheme.}
\label{fig:approach}
\end{figure*}

\section{Early Event-based Recognition}
\label{sec:eear}

Inspired by the benchmarking schemes for early RGB-based action recognition and anticipation, we introduce a framework to benchmark algorithms for early event-based recognition.\\
\textbf{Problem formulation.} Given an event stream $E = \{e_t\}^T_{t=0}$ of duration T and a class label $y$, the goal of early recognition is to produce a class prediction $p$ for a partial event stream $E_S = \{e_t\}^S_{t=0}$ where $S < T$. The class prediction is determined by the maximum index of a score function $f(E_s; \theta)$ with $\theta$ as a collection of parameters. With the extension to Top-K classification, the correctness $c$ is defined as:
\begin{equation}
c(E_S)^k = y \in argmax_k f(E_S; \theta) 
\end{equation}

If we evaluate the correctness $c$ for the iteration of $S$ from $0$ to $T$, we get a sequence of correctness scores $C^k = \{c^k_t \}^T_{t=0}$. The $\Delta t$ in this iteration can be seen as the output update rate, which might be different for a different score function $f$.\\
\textbf{Algorithmic benchmark.} Here we consider an idealized system in which the algorithm runtime is neglected. Hence, the correctness score $c(E_S)$ is available at time $S$. For datasets with a small number of classes, evaluating only for $k=1$ is enough. For datasets with more and similar classes, extending the evaluation to $k= [ 1, 3, 5 ]$ gives valuable insights and take-aways for approximate, early classification. We report the average $c(E_S)$ (accuracy) for the dataset's test split in tabular form at specific timestamps e.g. $S = [0.3s, 0.6s, 1.0s, 1.5s, 2.0s]$. Next to this, we plot at each sampled observation time $S$. For detailed analysis the evaluation can be plotted for every algorithmic update $\Delta t$. We name them \textit{accuracy-over-observation-time plots}.\\
Additionally, the number of synaptic operations (SynOps) can be added as a metric for computational efficiency. For SNN approaches which target neuromorphic hardware, the effective (non-zero) SynOps are added to the table or the plots. For detailed investigation, it is possible to also plot the accuracy over consumed SynOps (\textit{accuracy-over-synops plot}).\\
\textbf{System benchmark.} In this work we focus on algorithmic benchmarking. Nevertheless, we introduce a possible extension of the early recognition framework to system-level benchmarking. Therefore, the algorithmic runtime $t_f$ is also taken into account. This shifts the curves to the right in the \textit{accuracy-over-time plots}. Similarly, the metric for computational efficiency SynOps is replaced with the measured energy consumption which leads to \textit{accuracy-over-energy plots}.

\section{Spiking Two-Stream Net}
\label{sec:twosat}

In this section we present our spiking two-stream network. First we propose the general architecture in subsection~\ref{sec:twostream} and then deep-dive into the spiking and event-based units in \ref{sec:snu} and \ref{sec:egu}. In subsection~\ref{sec:optim}, we present a combined loss function and a readout neuron tailored for fast and accurate recognition.

\subsection{Trainable two-stream architecture}
\label{sec:twostream}

The concept of processing visual information with two concurrent paths is initially inspired by biological visual processing in the ventral and dorsal stream~\cite{two_stream_visual}. In deep learning and particularly for action recognition, two-path networks~\cite{TwoPathCNN, SlowFast, I3D} have been proposed. The same idea has also been adapted to SNNs and event-based action recognition~\cite{two_stream_snn}. Our two-stream architecture, visualized in Figure~\ref{fig:approach}, also consists of two streams which we call ventral and dorsal according to the biological example. The main novelty of our architecture is that it is fully trainable, with less initial biases on how the two-streams should behave. We put one common convolutional spiking neuron layer at the beginning and then split up into two streams (each with four convolutional spiking neuron layers). The ventral stream is configured with strides of 2 and growing channel width. The dorsal stream alternates between strides of 2 and 1, and slower growing channel width which is compressed by the final layer to the initial width. The spiking features from both streams are flattened, concatenated and processed by the fusion and classification layers.

\subsection{Convolutional spiking neuron layers}
\label{sec:snu}

The common layer and each layer of the ventral and dorsal streams consists of a 2D convolution, batch normalization and a spiking neuron unit. We define spiking activations as $s$ and the weighted and normalized neuron input as $x$ which leads to the following discrete time equation:
\begin{equation}
x[t] = BN(Conv(s[t])) 
\end{equation}
$x[t]$ is accumulated within the spiking neuron units according to their dynamics. In our work we compare parametric LIF (PLIF) and adaptive LIF (adLIF).\\
\textbf{PLIF.} The PLIF~\cite{plif} dynamic equation is a simple leaky integration with a trainable decay $\alpha_v$ per layer. This leads to the following membrane potential equation:
\begin{equation}
v[t] = \alpha_v v[t-1] + (1-\alpha_v) x[t]
\end{equation}
\textbf{adLIF.} Adaptive LIF neurons~\cite{adlif} involve a second state variable $w[t]$ and trainable parameter vectors $\alpha_v$, $\beta_w$, $a$ and $b$.  These allow to train the adLIF unit for complex behavior like spike frequency adaptation and membrane potential oscillation which allows for better sparsity and temporal processing~\cite{adlif}. All these parameters are optimized per channel, which leads to heterogeneous spiking neuron layers. The following equations describe the neuron behavior:
\begin{equation}
\begin{aligned}
& v[t] = \alpha_v v[t-1] + (1-\alpha_v) (-w[t-1] + x[t])\\
& w[t] = \beta_w w[t-1] + (1-\beta_w) (a u[t-1] + b s[t])
\end{aligned}
\end{equation}
Both spiking neuron units emit spikes if the membrane potential $v[t]$ crosses a threshold $\vartheta$:
\begin{equation}
s[t] = v[t] \geq \vartheta
\label{eq:02st}
\end{equation}
If a spike is emitted, the membrane potential is reset according to this equation:
\begin{equation}
v[t] = v[t] (1-s[t])
\label{eq:02rst}
\end{equation}

\subsection{Event-based gated fusion layer} % and ConvEGU
\label{sec:egu}
We flatten and concatenate the spiking activations from the ventral $s[t]_{ventral}$ and dorsal streams $s[t]_{dorsal}$ to $s[t]_{two-stream}$. These are processed by an event-based gated fusion layer. It consists either of an event-based gated recurrent unit (EGRU~\cite{egru}) or our proposed lightweight alternative: an event-based gated unit (EGU) inspired by the MinGRU\cite{mingru}.\\
\textbf{EGRU.} As proposed in \cite{egru} the equations contain an update gate $u[t]$, a reset gate $r[t]$, the candidate state $z[t]$ and the internal state $c[t]$. All gating signals are weighted functions of inputs $s[t]$ and state $c[t]$. An event generation mechanism is added in order to emit sparse, real-valued events $e[t]$.\\
\textbf{EGU.} Based on the EGRU concept and simplified MinGRUs~\cite{mingru}, we propose the event-based gated unit (EGU). As in \cite{mingru}, we remove the gate dependencies from the previous state as well as the reset gate $r[t]$, and drop the range restriction of $z[t]$. Doing this, we derive the following EGU equations for the update gate $u[t]$, the candidate state $z[t]$, the state $c[t]$, and events $e[t]$ with Heavyside function $H$ and threshold $\theta$:

\begin{equation}
\begin{aligned}
& u[t] = \sigma (Linear(s[t])) \\
& z[t] = Linear(s[t]) \\
& c[t] = u[t] c[t-1] + (1-u[t]) z[t] - e[t-1]\\
& e[t] = c[t] H(c[t] - \theta)
\end{aligned}
\end{equation}

Compared to the full EGRU, our EGU does not require matrix multiplications of weights and internal states and only needs 33\% of the EGRU parameters. Also, a future implementation on neuromorphic hardware like Loihi 2~\cite{loihi2} is simpler.

\subsection{Classification layer and loss}
\label{sec:optim}

We add a linear layer and non-spiking leaky-integrate neuron units (LI) for classification. The LI units have a trainable decay vector $\alpha_v$ per output class and the following dynamics:
\begin{equation}
v[t] = \alpha_v v[t-1] + (1-\alpha_v) Linear(e[t])
\end{equation}

%Loss: CE of average membrane potential (CEM), Average CE of regularly sampled membrane potential (TET)\\
Training the SNN is performed using a loss function which takes into account the LI membrane potential $v[t]$ and the target $y$. We propose two loss functions: the mean membrane potential cross entropy loss ($L_{CEM}$) and the average over N-times sampled membrane potential values cross entropy loss ($L_{TET}$~\cite{tet_loss}). Moreover, we also evaluate a combination of both loss functions $L_{Comb.}$ described by the following equations:
\begin{equation}
\begin{aligned}
& L_{CEM} = CE(\frac{1}{T} \sum_{t=0}^T v[t], y), L_{TET} = \frac{1}{N} \sum_{n=0}^N CE(v[n], y) \\
& L_{Comb.} = L_{CEM} + L_{TET}
\end{aligned}
\end{equation}

$L_{CEM}$ is designed to optimize for the best final accuracy over the complete sequence duration while $L_{TET}$ enforces also early prediction at every $n$-th timestep.

%Validation Readout: LI neuron to accumulate information + classify based on mean membrane potential (mean) or last membrane potential (last)\\
The prediction readout $x$ is also performed based on $v[t]$. Here, we compare using the mean potential $x = mean(v)$ or the last value $x = v[T]$. 
For implementation on neuromorphic hardware, we propose a \textbf{early recognition neuron} incorporating the LI dynamics, an early threshold on $v[t]$ and a second threshold on the running mean of $v[t]$. This allows for an early Top-K prediction and an accurate final classification.

%Spiking Readout Neuron: LI neuron to accumulate information + confidence threshold on membrane potential (early, coarse spikes) + running average over membrane potential and confidence threshold (slow, accurate spikes)\\

%% file: sec/4_experiments.tex
\section{Experiments}
\label{sec:experiments}

In this section we use the introduced early event-based recognition framework to evaluate the novel two-stream SNNs. First, we run a comparative study for the task of Top-1 early action recognition. Then, we expand to Top-K performance in order to study very early recognition. Finally, we showcase how we transferred our methods to a real-world use-case of triggering human motion capture for sports analytics.

\subsection{Dataset and Setup}
We use the THU EACT-50 dataset~\cite{thu_eact_50} throughout our experiments. This dataset consists of 10,500 recording of 105 subjects and contains 50 action classes. Compared to other event-based datasets of this size, it is directly recorded with an event camera, not converted or recorded from screen. The 50 actions range from full-body movements like \textit{running} over small hand gesture towards multi-player actions like \textit{shaking hands}. It also contains hard to distinguish action pairs (\textit{put on glasses} vs. \textit{put off glasses}) or triplets (\textit{sit down}, \textit{stand up} and \textit{sit and stand}).\\
The event stream samples have a duration of 2 to 5s. For the training, we randomly crop out 1000ms sequences from each sample. For early recognition evaluation, we start evaluating from the beginning of each recording. The spatial resolution of the raw event streams is 1280 x 800. For all our experiments, we center-crop to 600 x 600 and further downsample to 100 x 100. Temporally, the events are binned to frames according to a constant bin time parameter (e.g. 2ms). We sum up the events which fall into each downsampled bin per polarity. This simple encoding allows for event-by-event like processing and carries potential for implementation on asynchronously-operated neuromorphic hardware.\\
During the training we apply event-based augmentation strategies including random shift, zoom and horizontal flips as well as dropout after each spiking neuron layer. As in \cite{thu_eact_50}, we train all our models for 80 epochs.

\subsection{Comparative Study: Top-1 Performance}

%Accuracy over observation time plots for: all with SNN and 3D CNN baseline
%\begin{itemize}
%\item Choice of time bin
%\item Two path vs. single path networks
%\item PLIF, AdLIF and EGRU
%\item Optimizations (CE vs. TET, mean/last/max mem.)
%\end{itemize} 

\begin{figure}[h]
\centering
\includegraphics[clip,width=0.7\columnwidth]{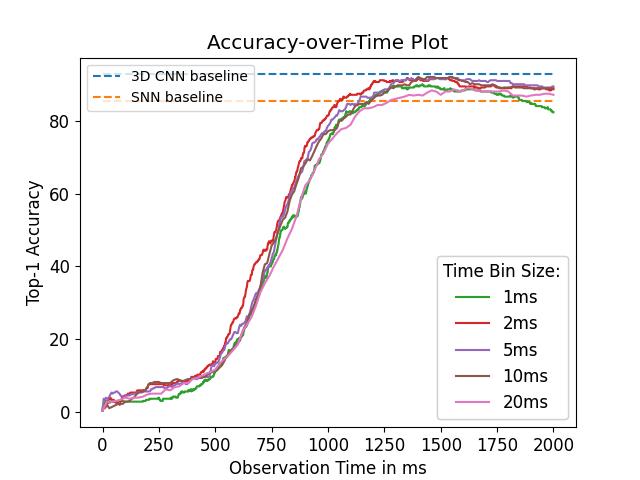}
\caption{Early recognition plots for PLIF+EGRU Two-Stream Net compared for different time bin sizes.}
\label{fig:plot_time_bin}
\end{figure}

We start the comparative study on early action recognition with two parametrization experiments. If not stated differently, we select the two-stream network with PLIF neurons, an EGRU layer for fusing the streams and a LI readout layer. By default, both loss and readout are based on the mean membrane potential. We provide the results in form of accuracy-over-observation-time plots and summarize them in the subsequent Table~\ref{tab:top-1}.\\
\textbf{Choice of time bin.} First we analyze the impact of the time bin duration on the early action recognition accuracy. In Figure~\ref{fig:plot_time_bin} time bin sizes from 1ms up to 20ms are compared.
%Discussion + \textbf{Take-away sentence}.\\
The red curve for a small time bin of 2ms generally performs best, especially for an observation time of below 1000ms. When decreasing the time bin size further to 1ms, the performance drops, same goes for increased time bin size. \textit{Take-away: Use a small time bin size of 2ms for subsequent experiments.}

\begin{figure}[h]
\centering
\includegraphics[clip,width=0.7\columnwidth]{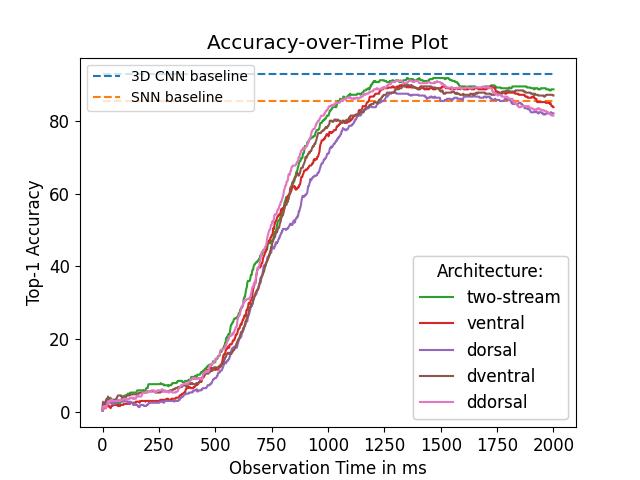}
\caption{Early recognition plots for PLIF+EGRU Two-Stream and Single-Stream networks. A "d" stands for doubled channel dimension.}
\label{fig:plot_snn_arch}
\end{figure}

\textbf{Choice of SNN architecture.} Next, we study the SNN architecture and in particular the impact of our proposed two-stream approach. Therefore, we compare the two-stream PLIF network with the single stream variants and single stream variants with doubled number of channels. The latter are included for a fairer comparison in terms of network capacity and compute operations. In Figure~\ref{fig:plot_snn_arch} we show the accuracy-over-observation-time plots.
%Discussion + \textbf{Take-away sentence}.\\
The two-stream SNN outperforms its single-path variants and also the ones with twice the channels for both early recognition and long-term classification. \textit{Take-away: Use two-stream SNN for further early action recognition experiments.}

\begin{figure}[h]
\centering
\includegraphics[clip,width=0.7\columnwidth]{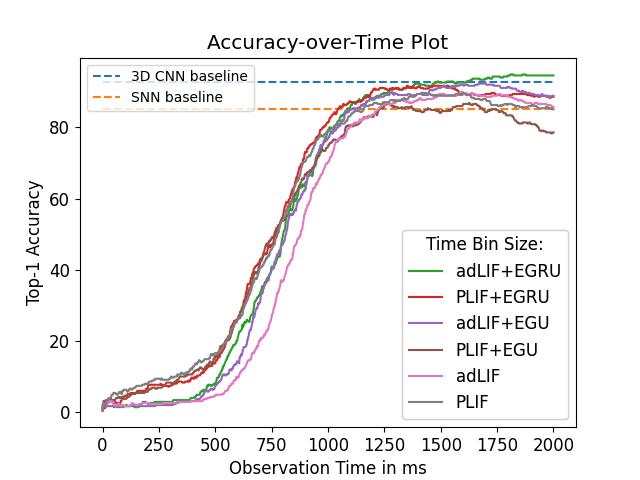}
\caption{Early recognition plots for Two-Stream SNNs with different spiking/event-based units.}
\label{fig:plot_neuron_units}
\end{figure}

\textbf{Impact of spiking/event-based neuron units.} Based on these two studies, we select a time bin duration of 2ms and the two-stream SNN architecture for further evaluation. Here we study the effect of spiking neuron units on early recognition performance. The two-stream feature extractor is evaluated for PLIF and adLIF neurons. For the fuse-streams layer we compare EGRU, EGU and PLIF/adLIF variants. The results are shown in Figure~\ref{fig:plot_neuron_units} and also in detail including additional metrics like parameter count and effective synaptic operations in Table~\ref{tab:top-1}.
%Discussion + \textbf{Take-away sentence}.\\
The PLIF networks perform better for early recognition, adLIF variants for long-term classification. Adding an event-based gated unit improves performance. EGRUs lead to higher accuracy, EGUs can save MAC operations and parameters (Table~\ref{tab:top-1}). Our two-path adLIF plus EGRU network even outperforms the previous SOTA 3D CNN on THU EACT-50 by about more than 2\% in accuracy and less than two orders of magnitude synaptic operations. \textit{The key take-aways are: Event-based gated units improve accuracy. adLIF networks are very sparse and good for long-term classification. PLIF networks are less sparse, but better for early recognition.}

\begin{table*}[ht!]
\centering
\begin{threeparttable}
\caption{Top-1 Performance evaluation on THU EACT-50.} 
\begin{tabular}{l|l|l||c|c|c|c|c||c|c|c} 
    %\hline
    %\multicolumn{2}{c||}{\multirow{2}{*}{Approach}} &  \multicolumn{5}{c||}{Top-1 Accuracy in \%} & \multirow{2}{*}{Params (M)} & \multicolumn{2}{c}{Eff. SynOPs} \\
    %\multicolumn{2}{l||}{}  & 0.3s & 0.6s & 1.0s & 1.5s & 2.0s & & MACs (G) & ACs (G) \\
    \multicolumn{3}{c||}{\textbf{Approach}} &  \multicolumn{5}{c||}{\textbf{Top-1 Accuracy in \%}} & \textbf{Params (M)} & \multicolumn{2}{c}{\textbf{Eff. SynOPs}\tnote{1}} \\
    Network & Loss & Readout  & 0.3s & 0.6s & 1.0s & 1.5s & 2.0s & (M) & MACs (G) & ACs (G) \\
    \hline \hline
    \multicolumn{3}{l||}{Motion SNN~\cite{motion_snn}} &
    - & - & - & - &  85.3\tnote{2} &
    - & - & - \\
    \multicolumn{3}{l||}{EV-ACT 3D CNN~\cite{thu_eact_50}} &
    - & - & - & - &  92.7\tnote{2} &
    21.3 & 29.0 & - \\
    \hline %\hline
    %Network & Loss/Readout &
    %\multicolumn{5}{c||}{} & \multirow{2}{*}{} & \multicolumn{2}{c}{} \\
    %\hline \hline
    Two-Path PLIF & CEM & Avg. Mem. &
    8.7 & 25.4 & 75.5 & 87.7 & 85.1 & 
    \textbf{1.31} & \textbf{0.004} & 0.23 \\
    Two-Path PLIF + EGU & CEM & Avg. Mem. &
    6.5 & 27.1 & 75.2 & 86.8 & 78.8 & 
    2.49 & 0.005 & 0.19 \\
    Two-Path PLIF + EGRU & CEM & Avg. Mem. &
    7.9 & 27.8 & 81.9 & 92.1 & 89.7 & 
    4.46 & 0.018 & 0.22 \\
    Two-Path adLIF & CEM & Avg. Mem. &
    2.4 & 9.9 & 67.4 & 88.7 & 86.4 & 
    1.32 & 0.004 & \textbf{0.040} \\
    Two-Path adLIF + EGU & CEM & Avg. Mem. &
    2.3 & 15.8 & 75.7 & 92.1 & 91.1 & 
    2.50 & 0.005 & 0.059 \\
    Two-Path adLIF + EGRU & CEM & Avg. Mem. &
    2.5 & 18.3 & 76.7 & \textbf{94.0} & \textbf{94.9} & 
    4.47 & 0.023 & 0.046 \\
    Two-Path PLIF + EGRU & TET & Avg. Mem. &
    23.7 & 35.5 & 69.7 & 87.5 & 87.8 & 
    4.46 & 0.015 & 0.43 \\
    Two-Path PLIF + EGRU & TET & Last Mem. &
    \textbf{25.2} & \textbf{46.1} & 78.7 & 79.0 & 66.9 & 
    4.46 & 0.015 & 0.43 \\
    Two-Path PLIF + EGRU & Comb. & Avg. Mem. &
    18.4 & 34.4 & 75.1 & 89.3 & 88.9 & 
    4.46 & 0.014 & 0.32 \\
    Two-Path PLIF + EGRU & Comb. & Last Mem. &
    17.7 & 43.4 & \textbf{82.6} & 79.7 & 59.4 & 
    4.46 & 0.014 & 0.32 \\
    
\end{tabular}
\begin{tablenotes}
\item[1] Effective synaptic operations are reported for 2.0s.
\item[2] Accuracy reported for the complete samples.
\end{tablenotes}
\label{tab:top-1}
\end{threeparttable}
\end{table*}

\begin{figure}[h]
\centering
\includegraphics[clip,width=0.7\columnwidth]{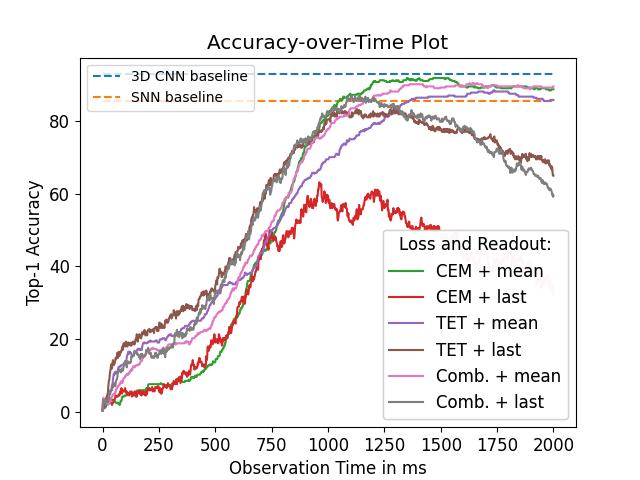}
\caption{Early recognition plots for Two-Stream SNNs and different loss/readout.}
\label{fig:plot_loss}
\end{figure}

\textbf{Impact of loss/readout.}
In Figure~\ref{fig:plot_loss} we compare the different loss and readout choices. We add the compared approaches also to Table~\ref{tab:top-1}. 
%Discussion + \textbf{Take-away sentence}.\\
The variants trained with TET loss show improved early recognition. Reading out the last membrane potential value also slightly better early recognition, but the accuracy drops for longer evaluation compared to mean readout. \textit{Our combined loss implements a trade-off and performs well for both early last and long-term mean readout.}

%Comparison with SotA in Table: Approach, Accuracy for 0.5s, 1s, 1.5s, 2s, best, (No. of Params, Eff. MACs/ACs for best) 
%\begin{itemize}
%\item THU EACT-50: SlowFast 3D CNN, their SNNs, Ours
%\item DVSGesture: 
%\item THU EACT-50-CHL: SlowFast 3D CNN, their SNNs, Ours
%\end{itemize} 

\subsection{Early Top-K Performance}

\begin{figure}[h]
\centering
\includegraphics[clip,width=0.7\columnwidth]{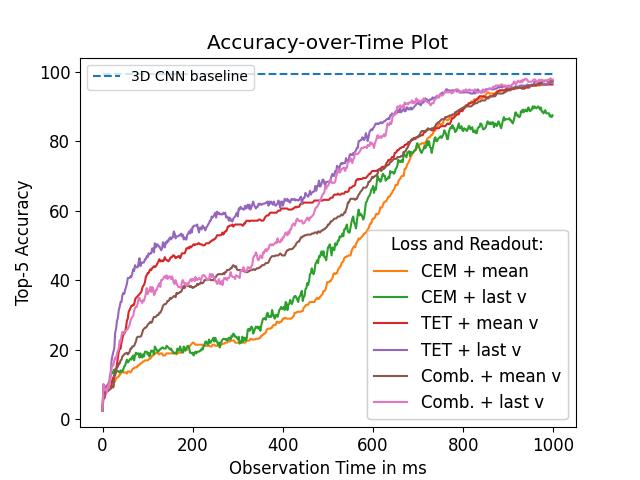}
\caption{Top-5 early recognition plots for Two-Stream SNNs and different loss/readout choices.}
\label{fig:plot_top5}
\end{figure}

%Early Top-5 Action Recognition.\\
Next to early Top-1 evaluation, we also benchmark our two-stream PLIF plus EGRU models for approximate early classification. This evaluation is done via the Top-5 accuracy. In Figure~\ref{fig:plot_top5} the curves are plotted. The best early recognition model allows for 50\% Top-5 accuracy already within 100ms of observation. Leveraging this approximate early recognition capabilities highlights the advantage of event-based vision sensors and high-rate SNNs compared to frame-based processing approaches where only 2-3 frames were captured within the same time.

%Maybe qualitative evaluation: membrane potential for plif vs. adlif comparison
%\begin{itemize}
%\item Running, running circles, staggering
%\item Reverse actions: Put glasses on, take glasses off (31-39)
%\item Sequentially-combined actions: Sit down, stand up, sit down \& stand up (13, 14, 15)
%\end{itemize}

%\subsection{Streaming Benchmark}

%Accuracy over latency plots on Nvidia Orin Nano (or RTX 2080):
%Ours vs. 3D CNN vs. Reference CNN+GRU

\subsection{Transfer to real-world human motion capture triggering}

\begin{figure}[h]
\centering
\includegraphics[clip,width=\columnwidth]{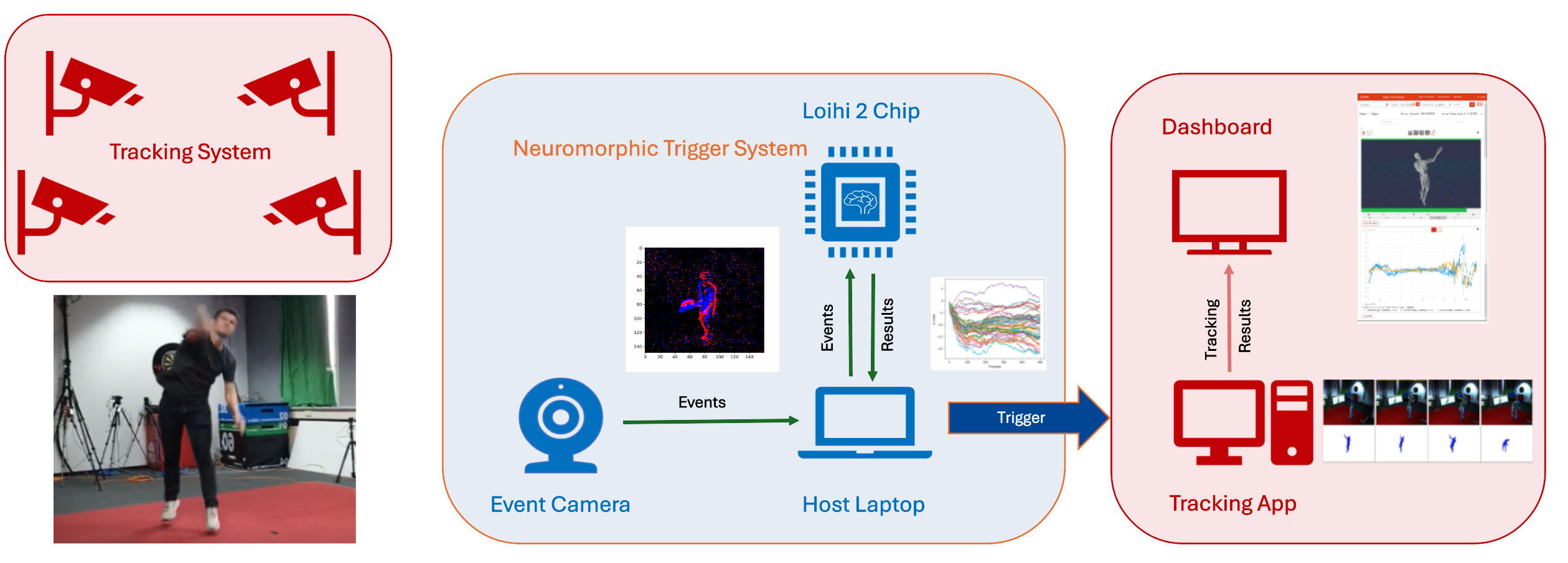}
\caption{Block diagram of our neuromorphic trigger system running our two-stream SNNs on a neuromorphic processor and its interaction with the human motion capture system.}
\label{fig:blockd_spikingbody}
\end{figure}

\begin{figure}[h]
\centering
\begin{subfigure}[b]{0.48\columnwidth}
        \centering
        \includegraphics[width=\textwidth]{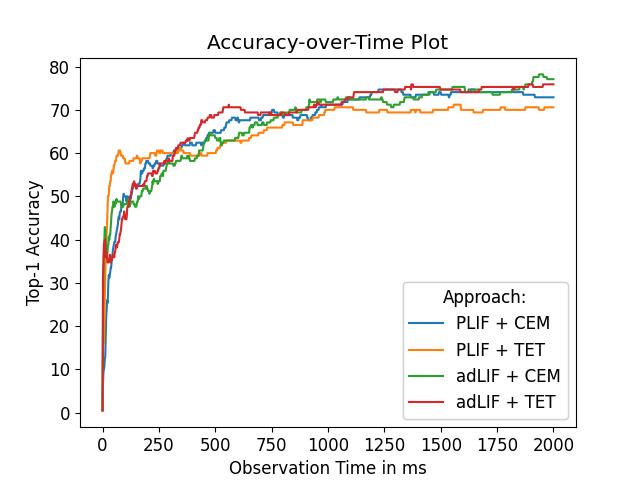} 
        \caption{Top-1 Accuracy} \label{fig:sub1}
    \end{subfigure}
    \hfill
    \begin{subfigure}[b]{0.48\columnwidth}
        \centering
        \includegraphics[width=\textwidth]{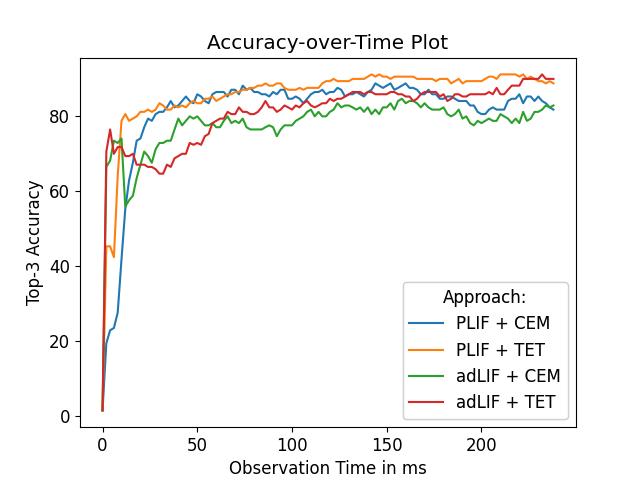} 
        \caption{Early Top-3 Accuracy} \label{fig:sub2}
    \end{subfigure}
\caption{Early recognition plots for evaluation with SpikingBody Dataset.}
\label{fig:plot_spikingbody}
\end{figure}

Finally, we also transfer the acquired knowledge to a real-world application in sports. \\
\textbf{Use-Case.} Human motion capture systems are highly used for health and sports analytics~\cite{mocap_survey}. Within recent years, markerless motion capture systems based on multiple stationary cameras became commercially available. But running them in always-on mode requires a lot of power and produces GBs of data~\cite{mocap_survey}. The current alternative is to have a human operator manually starting the capture. We propose to leverage early event-based action recognition as a trigger to automatically and data-dependently start recording. In Figure~\ref{fig:blockd_spikingbody} we show the block diagram of the final neuromorphic trigger system and its potential assets. \\
\textbf{Setup.} In this work, we focus on the SNN network which will be put on a neuromorphic processor in the next step. Nevertheless, we can already evaluate it algorithmically based on our proposed framework. The specific use-case is tracking of tennis serves. Hence, we recorded the \textit{SpikingBody Dataset} consisting of tennis serves, other tennis action (backhand, forehand) and noise actions (running, jumping, arm and leg movements) with a Prophesee event camera. We recorded 12 subjects which leads to around 1100 train and 400 test event streams of 1.0s duration. We train the models for 50 epochs on the dataset.\\
\textbf{Evaluation.} In Figure~\ref{fig:plot_spikingbody} we show the accuracy-over-time plot for different models on the tennis use-case. In the left Subfigure~\ref{fig:sub1} the Top-1 performance for mean membrane potential readout networks is displayed. As for the THU EACT-50 dataset the adLIF variants lead to better final classification accuracy up to 78\%. In the right Subfigure~\ref{fig:sub2} we visualize the Top-3 performance within the first 200ms of observation for last membrane potential readout. Already within the first 50 ms most approaches reach 80\% here which indicates a good coarse early action prediction.

%% file: sec/5_conclusion.tex
\section{Conclusion}
\label{sec:conclusion}
In this work we introduce an early event-based recognition evaluation framework which highlights the relevance of high-rate spiking processing. We propose a spiking two-stream net which outperforms the former state of the art on the THU EACT-50 dataset in final accuracy by 2\% while requiring only 20\% of parameters and more than two orders of magnitude less synaptic operations. By evaluating our approach for early Top-5 accuracy, we showcase the early prediction capabilities which we also transfer to the real-world use-case of triggering a human motion capture system for sports analysis. Our work acts as a foundation for extended research on early event-based recognition/anticipation and implementation of our model on neuromorphic hardware.